\documentclass{article}

\usepackage[final, nonatbib]{neurips_creativity_2020}

\usepackage[utf8]{inputenc} 
\usepackage[T1]{fontenc}    
\usepackage{url}            
\usepackage{booktabs}       
\usepackage{amsfonts}       
\usepackage{nicefrac}       
\usepackage{microtype}      

\usepackage{amsmath}

\usepackage{graphicx}
\usepackage{subcaption}
\usepackage{caption}
\usepackage{xcolor}
\usepackage{wrapfig}

\usepackage{algorithm}
\usepackage[noend]{algpseudocode}

\algrenewcommand\algorithmicrequire{\textbf{Precondition:}}
\algrenewcommand\algorithmicensure{\textbf{Postcondition:}}

\usepackage{color}
\usepackage{xcolor,colortbl}
\definecolor{matlab-blue}{rgb}{0,0.4470,0.7410}
\definecolor{matlab-orange}{rgb}{0.8500,0.3250,0.0980}
\definecolor{matlab-yellow}{rgb}{0.9290,0.6940,0.1250}
\definecolor{matlab-green}{rgb}{0.4660,0.6740,0.1880}
\definecolor{matlab-red}{rgb}{0.6350,0.0780,0.1840}
\definecolor{matlab-purple}{rgb}{0.4901,0.1803,0.5529}
\definecolor{ourmethod}{gray}{0.93}
\definecolor{mydarkblue}{rgb}{0,0.08,0.45}

\newcommand{\etal}{\textit{et al.}}

\newcommand{\comment}[1]{\iffalse#1\fi}

\usepackage[numbers]{natbib}
\usepackage[colorlinks]{hyperref}
\hypersetup{
  pdftitle={},
  pdfauthor={},
  pdfkeywords={},
  pdfborder=0 0 0,
  pdfpagemode=UseNone,
  colorlinks=true,
  linkcolor=mydarkblue,
  citecolor=mydarkblue,
  filecolor=mydarkblue,
  urlcolor=mydarkblue,
  pdfview=FitH,
}

\title{Spatial Assembly:\\Generative Architecture With\\Reinforcement Learning,\\Self Play and Tree Search}

\author{%
  Panagiotis Tigas \\
  University of Oxford\\
  \texttt{ptigas@robots.ox.ac.uk}
  \And
  Tyson Hosmer \\
  Bartlett School of Architecture, UCL\\
  \texttt{tyson.hosmer@gmail.com}
}

\begin{document}

\maketitle

\begin{abstract}
  With this work, we investigate the use of \emph{Reinforcement Learning} (RL) for generation of spatial assemblies, by combining ideas from Procedural Generation algorithms (\emph{Wave Function Collapse} algorithm (WFC) \cite{mxgmn}) and RL for Game Solving\comment{(\emph{Monte Carlo Tree Search \cite{chaslot2008monte}} (MCTS))}. WFC is a Generative Design algorithm, inspired by Constraint Solving \cite{Karth2017}. In WFC, one defines a set of tiles/blocks and constraints and the algorithm generates an assembly that satisfies these constraints. Casting the problem of generation of spatial assemblies as a \emph{Markov Decision Process} whose states transitions are defined by WFC, we propose an algorithm that uses \emph{Reinforcement Learning} and \emph{Self-Play} to learn a policy that generates assemblies which maximize objectives set by the designer. Finally, we demonstrate the use of our Spatial Assembly algorithm in Architecture Design.
\end{abstract}

\vspace{-.8em}
\section{Introduction}
\vspace{-1em}

We present a novel application of Deep Reinforcement Learning, coupled with a bespoke Constraint Solving algorithm for learning to generate spatial assemblies.


Constraint Satisfaction Problems (CSPs) consist of a finite set of rules and objects, whose composition/combination must satisfy a number of constraints \cite{apt2003principles}. CSP solvers have been effective in computation logic problems across many domains including decision making, game development, logic puzzles \cite{miguel2012dynamic,modi2001dynamic,simonis2005sudoku}. Design innovation through constraint solving has been extensively explored by Killian \etal~\cite{kilian2005design}, whose research has focused on constraints in design exploration and specifically bidirectional constraint solving methods \cite{kilian2005design}. Our approach explores building design as a multi-objective CSP, trained to evaluate each local decision based on the current state of the assembly to effectively negotiate evolving socioeconomic and environmental goals.

We begin by modeling architecture design as a CSP, extending the approach of Texture Synthesis and Model Synthesis \cite{merrell2007example}, and Wave Function Collapse \cite{mxgmn, Karth2017}, primarily applied to image-based procedural content creation and modeling in gaming. The algorithm extracts features and their relations from images and attempts to recreate similar distributions of those features procedurally creating images that
resemble a prototypical image.




However, using such algorithms to generate assemblies \footnote{assemblies, designs, and structures will be used interchangeably}  that optimize certain criteria, additionally to the constraints solving, is a difficult task because of the lack of differentiability and their dependency on black-box methods.
To elevate this limitation, we equip the search space of possible assemblies with an efficient learnable search operator/policy $\pi(a|s)$.


\begin{figure}
  \vspace{-2em}
  \centering
  \includegraphics[width=1.0\textwidth]{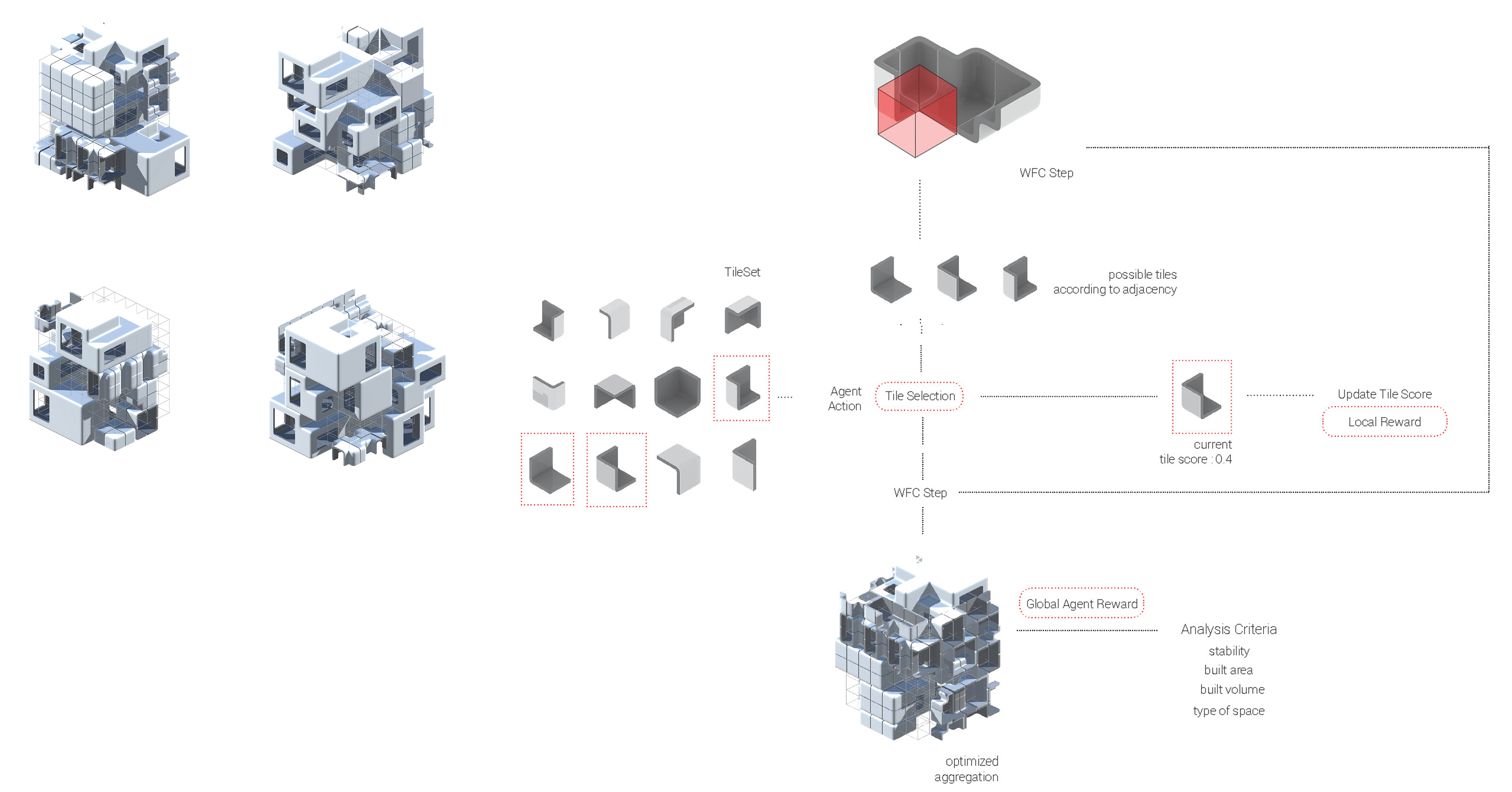}
  \caption{Spatial Assembly process.}
  \vspace{-1em}
\end{figure}

\vspace{-.8em}
\section{Algorithm}
\vspace{-1em}

First, we define a set of geometric tiles that form a dictionary $D$ of building blocks ($D_i$ represents the $i$-th tile of the set). Next, we set a rule of constraints, $C$, according to which these tiles can be combined. The problem then becomes to sequentially combine the tiles in order to create structures that are valid (no constraint is invalidated) and maximally cover the available canvas.

Wave Function Collapse starts with an empty state and selects an initial tile at random. Meanwhile, for each possible expansion node (expansion node is a connection point of a tile which is free) it keeps track of the number of tiles that can be connected which do not invalidate the constraints, termed entropy. WFC works by selecting the node with the least degrees of freedom (most constraint node) and expanding the node by randomly selecting a tile that satisfies the constraints. We can see the problem of generation of an assembly as solving a Markov Decision Process (MDP), where the state transitions are defined by WFC algorithm, actions are the tiles from the dictionary $D$, and rewards are defined according to the designer's goals.

\begin{algorithm}[H]
  \caption{Spatial Assembly - Rollout
      \label{alg:MCWFC}}
  \begin{algorithmic}[1]
      \State $S \leftarrow \emptyset$ \Comment{Initialize empty}
      \While{structure not complete or invalid}
      \State node = SelectNode($S$)\Comment{Select the most constraint node}
      \State tiles = GetValidTiles(node, $C$, $D$) \Comment{Get the set of tiles that satisfy the constraints}
      \State Sample $T_{\text{new}}$ from policy $\pi(a|S, \text{tiles})$
      \State Update $S$ by connecting the new tile $T_{\text{new}}$ at node
      \EndWhile
  \end{algorithmic}
\end{algorithm}

Spatial Assembly algorithm, replaces the random selection of the tiles with the policy $\pi(a|s)$, which returns a distribution over the available tiles (action $a$) according to their potential to maximize the future expected reward. We learn the policy with Proximal Policy Optimization \cite{schulman2017proximal}, a Reinforcement Learning algorithm which has enjoyed success in various domains of Artificial Intelligence. The complete rollout algorithm can be found in alg.~\ref{alg:MCWFC}.

Training the system occurs as follows. We start generating rollouts with an initially untrained policy until we reach a terminal state. We evaluate the terminal state according to the success and reward accordingly. For example, one reward signal we used was the maximum displacement observed on the final structure after it got simulated by the physics engine of Unity3D (reward capturing the structural stability of the assembly). We then use Proximal Policy Optimization to update the value function and the policy for the next round. We let the system self-play until convergence. This approach can be seen as Policy Gradient Search \cite{anthony2019policy}.

\section{Acknowledgements}

The authors would like to thank Dave Reeves, Octavian Gheorghiu, and Ziming He, design masters and tutors at Living Architecture Lab, The Bartlett School of Architecture, and the students Elahe Arab, Barış Erdinçer, Yifei Jia, Georgia Kolokoudia (IRSILA project, 2020 cohort), Athina Athiana, Evangelia Triantafylla, Ming Liu (NOMAS project, 2019 cohort), Jelena Peljevic, Yekta Tehrani, Shahrzad Fereidouni, Noura Alkhaja (ArchiGO project, 2018 cohort).
\bibliography{references}
\bibliographystyle{abbrvnat}

\appendix

\section{Application: Spatial Assembly in Architecture Design}
\vspace{-1em}

This methodology was applied in three design projects, ArchiGo(2018), Nomas(2019), and ISIRLA(2020), at Bartlett School of Architecture, Living Architecture Lab.

The ArchiGo (fig.~\ref{fig:archigo}) project was developed by iteratively designing and testing many spatial part sets with different characteristics and relations evaluated for their ability to avoid contradictions and meet user-defined spatial objectives (Figure 2).



\begin{figure}[H]
  \centering
  \includegraphics[width=0.7\textwidth]{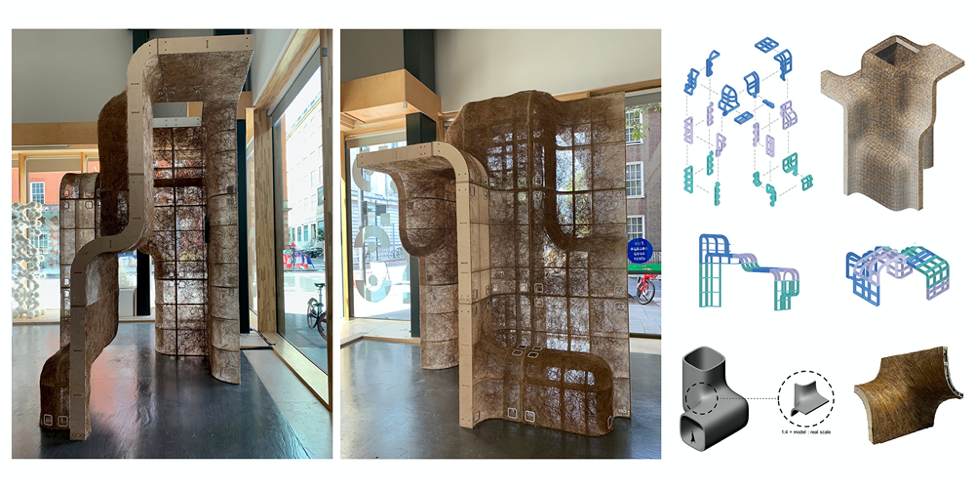}
  \caption{NoMAS Project (2018)}
  \label{fig:nomas}
\end{figure}

\begin{figure}[H]
  \centering
  \includegraphics[width=0.7\textwidth]{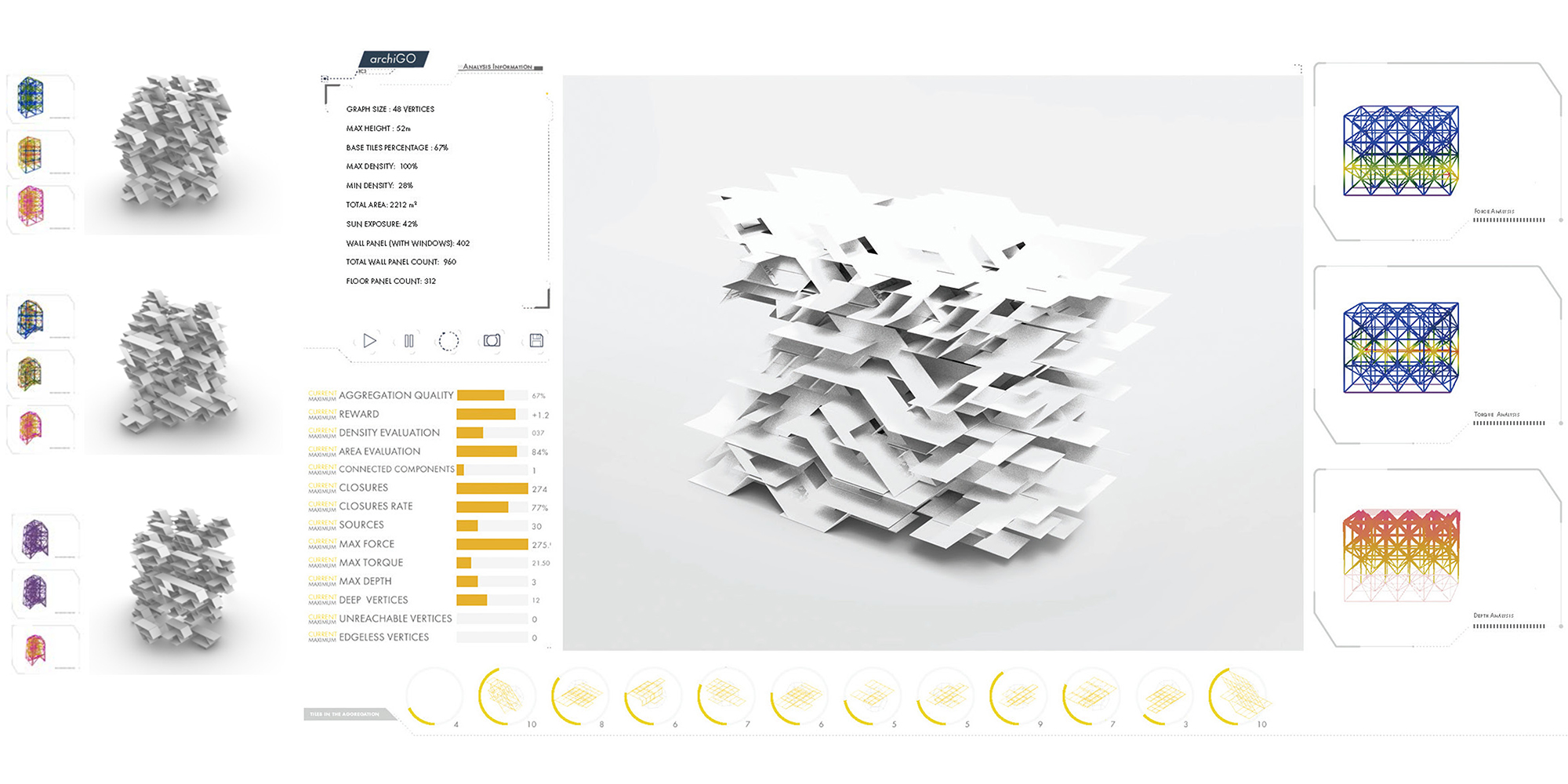}
  \caption{ArchiGo Project (2019)}
  \label{fig:archigo}
\end{figure}

In the NOMAS project (fig.~\ref{fig:nomas}), we investigate the potential for this method to re-think housing strategies and invent new spatial languages composed of simple prefabricated parts. The strategy is demonstrated through the digital process applied to the physical production of a 3.5-meter-tall spatial prototype assembled with human labor from coconut fiber composite parts.

\begin{figure}[H]
  \centering
  \includegraphics[width=0.7\textwidth]{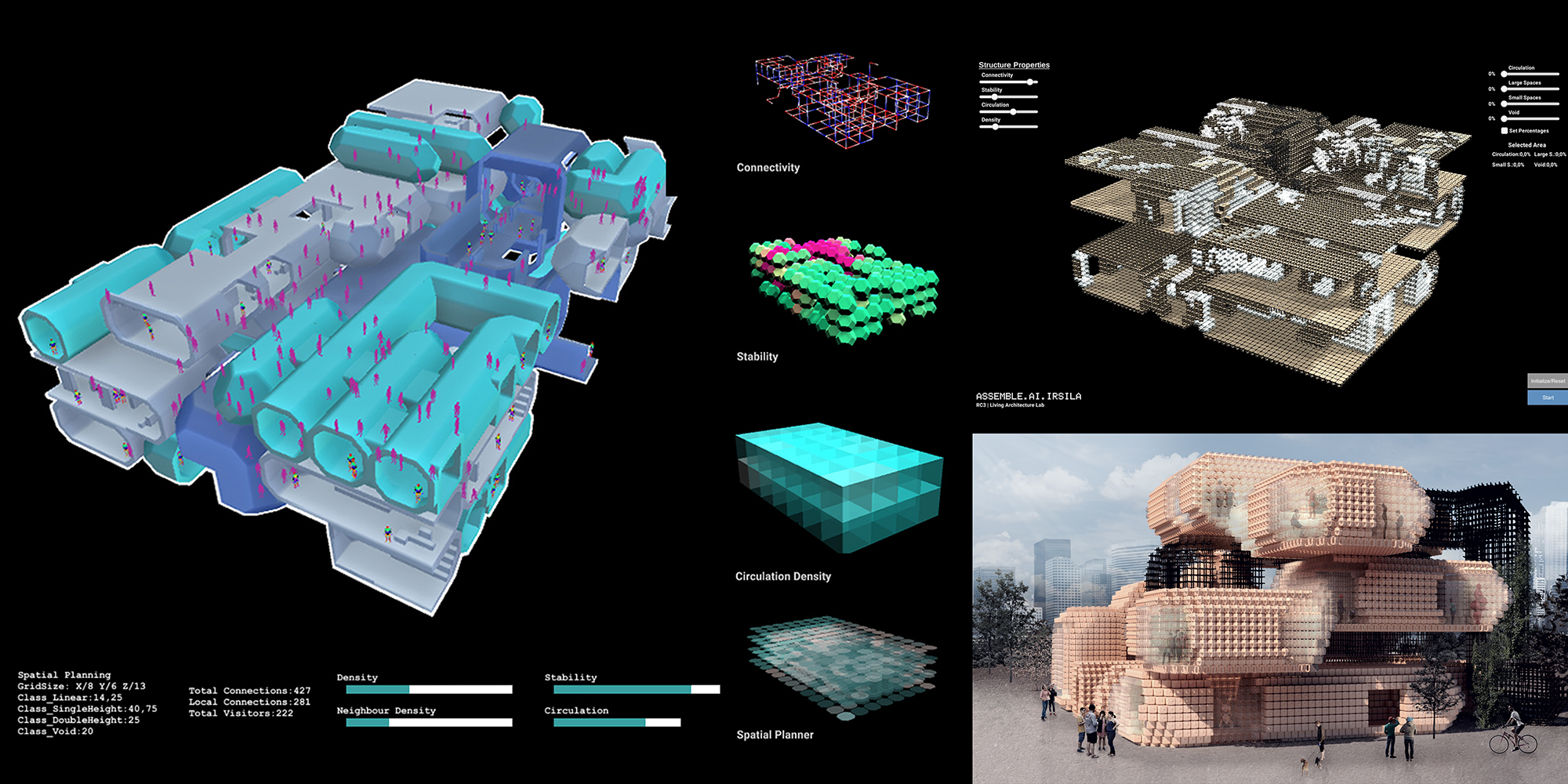}
  \caption{IRSILA Project (2020)}
  \label{fig:irsila}
\end{figure}

IRSILA (fig.~\ref{fig:irsila}) applies the methodology to a reconfigurable cultural center where spatial parts are constructed from smaller prefabricated units assembled and reconfigured by autonomous distributed robots. Both demonstrate the potential for buildings with reconfigurable and adaptive life cycles.

\end{document}